\documentclass[conference]{IEEEtran}
\IEEEoverridecommandlockouts
\usepackage[compress]{cite}
\usepackage{amsmath,amssymb,amsfonts}
\usepackage{algorithmic}
\usepackage{graphicx}
\usepackage{textcomp}
\usepackage{xcolor}
 
\usepackage{booktabs}       
\usepackage{subfigure}

\usepackage{algorithm}
\usepackage{url}
\usepackage{flushend}

\def\BibTeX{{\rm B\kern-.05em{\sc i\kern-.025em b}\kern-.08em
    T\kern-.1667em\lower.7ex\hbox{E}\kern-.125emX}}
\begin{document}

\title{BRIEF: Backward Reduction of CNNs \\ with Information Flow Analysis}



\author{\IEEEauthorblockN{Yu-Hsun Lin}
\IEEEauthorblockA{HTC Research \& Healthcare \\
lyman\_lin@htc.com}
\and
\IEEEauthorblockN{Chun-Nan Chou}
\IEEEauthorblockA{HTC Research \& Healthcare \\
	jason.cn\_chou@htc.com}
\and
\IEEEauthorblockN{Edward Y. Chang}
\IEEEauthorblockA{HTC Research \& Healthcare \\
	edward\_chang@htc.com}
}

\maketitle

\begin{abstract}
This paper proposes {\bf BRIEF}, a backward reduction algorithm that 
explores compact CNN-model designs from the information flow perspective.
This algorithm can remove substantial non-zero weighting parameters (redundant neural channels) of a network by considering its dynamic behavior, 
which traditional model-compaction techniques cannot achieve.
With the aid of our proposed algorithm, we achieve significant model reduction on ResNet-$34$ in the ImageNet scale ($32.3\%$ reduction), which is $3 \times$ better than the previous result ($10.8\%$).
Even for highly optimized models such as SqueezeNet and MobileNet, we can achieve additional $10.81\%$ and $37.56\%$ reduction, respectively, with negligible performance degradation.
\end{abstract}

\section{Introduction}

Since the breakthrough performance demonstrated by convolutional neural networks (CNNs) on ImageNet, deep architecture has been successfully applied to a number of areas such as speech recognition, object tracking, and image classification.  
As the width and depth of a CNN is increased to improve prediction accuracy, the model complexity and training time increase as well.  Whereas model training can be sped up by employing a large number of GPUs, inferencing on mobile and wearable devices (e.g., mobile VR) faces the resource limitations of memory, power and computation. 
In this work, we utilize information flow analysis to perform CNN model reduction while preserving prediction accuracy.  

Traditionally, a complex CNN is simplified for embedded systems by using the {\em teacher-student} model \cite{fitnet,distill}.
Such simplification demonstrates that important properties of a CNN can be preserved when its model complexity is reduced. 
However, almost all model-reduction approaches treat a CNN as a black-box and simply compress the model parameters obtained by the training process. 
Much effort has recently been devoted to opening the black-box to better understand and interpret CNNs.  
The work by \cite{naftali-information} proposes the use of information theory to analyze the internal behaviors of a deep architecture. 
Inspired by this approach, our work incorporates information density to conduct model reduction.
Our information-based approach works orthogonally to the traditional black-box approach (discussed further in Section II), and can achieve additional compaction 
while preserving prediction accuracy.

We first conduct lesion studies to probe the dynamic nature of the network robustness in CNNs with the information density consideration.
We define \textit{convolution macroblock} as the set of convolution layers whose output feature maps have the same height and width.
Our lesion studies provide important clues for us to construct a hypothetical information flow structure. The hypothetical structure is useful for guiding us to design an effective model reduction algorithm. The hypothetical framework formulates a CNN as an information pipeline consisting of cascaded convolution macroblocks.
Our aim is to identify the fewest channel numbers\footnote{The input to a convolution layer is a set of input tensors, each of which is called a channel \cite{survey-efficient-dnn}.} with sufficient information flow between macroblocks so as to reduce their associated parameters.
We propose our \textit{backward reduction} algorithm named BRIEF, which incorporates the hypothetical information flow structure to achieve our twin design goals of model reduction and accuracy preservation.

The contributions of this work can be summarized as follows:
\begin{itemize}
	\item {\em Applicable to various CNN models.}
	We propose an information-based model reduction algorithm, {\em backward reduction}, which can be applied to any already compact CNN structures such as MobileNet \cite{mobilenet}, ResNet \cite{resnet}, and SqueezeNet \cite{squeezenet} to achieve further model-size reduction.
	
	\item {\em Significant reduction results.}
	With the aid of BRIEF, we are able to compress MobileNet to a model size that is smaller than that of SqueezeNet while achieving higher prediction accuracy.
	We also achieve a $32.3\%$ model reduction on ResNet-$34$ with ImageNet, which is $3 \times$ better than the state-of-the-art approach ($10.8\%$) presented in \cite{prune-filter}.
	Even in the case of SqueezeNet, a highly optimized model, we can achieve additional $10.8\%$ reduction with negligible prediction degradation.
    
    \item {\em Further model compaction.}
    The Convolution-layer dominant models are inherently compact and become the default choices for mobile devices. Our proposed framework can further shrink these highly optimized CNN models (e.g., MobileNet\cite{mobilenet}).
Most traditional compaction works regard the CNN as a black-box and perform either parameter elimination or compression within a limited range.
Our method utilizes the distribution trends of the information density and the dynamic nature of CNNs, which works orthogonally to the black-box approach.
Therefore, we can reduce the model size of CNNs further.
\end{itemize}

\begin{figure*}[t]
	\begin{center}
		\includegraphics[height=3cm]{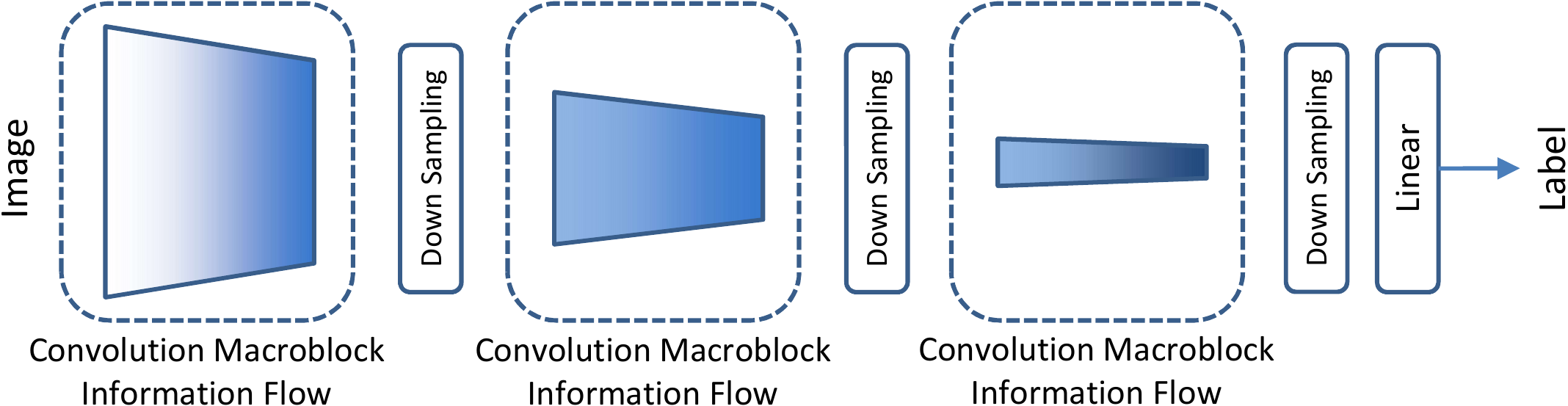}
		\caption{(a)The information flow structure conjectured from the the DPI theorem. The information density within the information flow is in a descending trend for each convolution macroblock. A  macroblock represents the set of convolution layers whose the output feature maps have the same height and width.}
		\label{fig-information-flow}
	\end{center}
\end{figure*}

To the best of our knowledge, this is the first work to achieve significant reduction on already highly compact CNN models using information flow analysis.
The remainder of this paper is organized into four sections.
Section II describes related work in CNN models and model-reduction techniques.
Section III explains our hypothetical information flow structure and presents
our proposed backward reduction algorithm.
Section IV details experiments. 
We offer our concluding remarks in Section V.

\section{Related Work}
The rapid increase of the number of parameters from the cascaded fully connected (FC) layers is the main reason for excessive CNN model size.
Recent developed CNN models (e.g., ResNet \cite{resnet}, DenseNet \cite{densenet}, and MobileNet \cite{mobilenet}) show 
that with the aid of improved structural design, the cascaded FC layer design is not mandatory for achieving high prediction accuracy.
These improved designs include, but are not limited to, batch normalization \cite{batch-normalization} and bottleneck structure \cite{googlenet}.
Though these new models without many cascaded FC layers are highly compact, we propose BRIEF from the information density perspective to further compact these models.
 
This section first reviews existing
CNN models (Section~\ref{sec-cnn-categories}) and then traditional 
black-box model-reduction approaches (Section~\ref{sec-reductions}).
In comparison with the traditional methods, our BRIEF is a coarse-grained model compaction technique applicable to both FC-layer dominant and Convolution-layer dominant CNN models.

\subsection{Categories of CNNs}
\label{sec-cnn-categories}

A CNN typically consists of convolutional layers, pooling layers, and FC layers.
Based on the distribution of the model parameters in a CNN, we can categorize the CNN as either FC-layer dominant or Convolution-layer dominant.

\begin{itemize}
	\item {\em FC-layer dominant models}. A model is considered as FC-layer dominant when the parameters of FC layers comprise more than $50\%$ of the total parameters in a CNN. 
	The well-known AlexNet \cite{alexnet} and VGGNet \cite{vgg} models are examples of FC-layer dominant CNNs.
	CNNs in this category usually possess huge model footprints that are mainly contributed by the parameters of the FC layers.
    The parameters of the FC layers in VGG-$11$ account for $472$MB ($93\%$) out of the total $507$MB storage.
	
	\item {\em Convolution-layer dominant models}.
	The recent trend of CNNs replaces the cascaded FC layers with a global average pooling layer.
	The parameters of FC layers now comprise less than $20\%$ of parameters in the whole model. 
	CNNs in this category are compact models with high prediction accuracy. For example, MobileNet takes up only $16.25$MB of storage but possesses VGG-level accuracy.
	
\end{itemize}

\subsection{Model Reduction Techniques}
\label{sec-reductions}

Based on the level of involved structures, the model reduction techniques can be classified into two categories: fine-grained and coarse-grained~\cite{survey-efficient-dnn}.

\begin{itemize}
	\item {\em Fine-grained approaches.}	
	There are various reduction techniques applied on the filter/kernel level. Such techniques include sparse convolution \cite{sparse-conv} and deep compression \cite{deep-compression}.
	The irregular structures introduced on the filter/kernel level often require special hardware acceleration \cite{eie}.	
    Replacing the cascading FC layers with an average pooling layer introduces negligible performance degradation.
    FC layer pruning \cite{net-trim} usually achieves significant reduction since FC layers are sparse in nature.    	
	Binary filter/kernel approximation methods are also popular algorithms to model reduction \cite{binarynet,xnor-net,quantized-cnn,fixed-point-network}. 
	However, retaining the prediction accuracy of a binary CNN is a challenging issue. This issue is addressed and relaxed by using multiple binary representations \cite{aaai-accurate-binarynet,abcnet}.		

	\item {\em Coarse-grained approaches.}	
	The coarse-grained techniques conduct filter pruning \cite{prune-filter,network-slim,iccv-prune-channel}, which removes irrelevant filters/kernels from the model.
    The filter importance is determined by filter weight magnitude \cite{prune-filter} or derived from the similarity among inter-layer feature maps \cite{iccv-prune-channel}.
    The work of \cite{network-slim} integrates an additional scaling factor for each filter output during the training process, which the magnitude of the factor reflects the filter importance.
    Conducting pruning at the filter-level can free us from designing kernel-specific hardware/software solutions.
	
\end{itemize}

\begin{figure*}[t]
	\begin{center}
		\centerline{\includegraphics[width=0.8\textwidth]{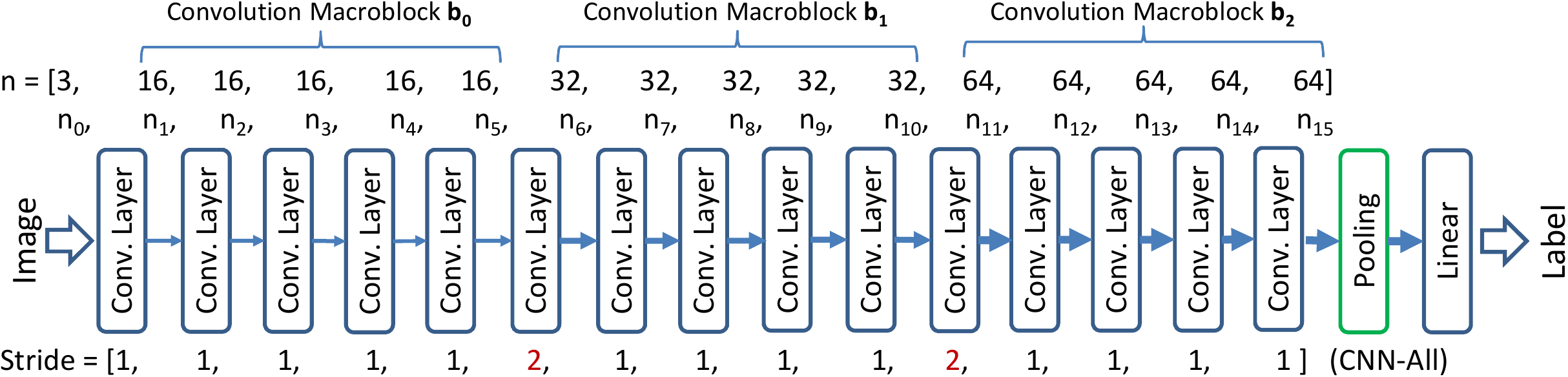}}
		\caption{A sequential CNN example where the convolution depth = {$15$}. Each  layer is composed of Conv-BN-Relu modules.}
		\label{fig-cnn}
	\end{center}
\end{figure*}

\section{Method}

In this section, we study the information flow of CNNs using the data processing inequality of information theory \cite{information-theory}.
We investigate the relationship between the network robustness 
and the information density of CNNs by our designed lesion studies.
Our lesion studies reveal the dynamic nature of the network robustness of CNNs for dealing with the distortions caused by channel removal.
We utilize the insights observed from our lesion studies to later (in Section~\ref{sec-proposal})
propose our backward model-reduction algorithm.

\subsection{Descending Trend of Information Density}

The data processing inequality (DPI) theorem \cite{information-theory} shows that for the variables forming a Markov chain $X \rightarrow X' \rightarrow Y$, we have
\begin{equation}\label{eq-dpi}
I(X; X') \geq I(X; Y),
\end{equation}
where $I(X;Y)$ represents the mutual information between variables $X$ and $Y$.

Let us consider a CNN as an information pipeline where its input is images and post-processing operations are convolution and filtering. Based on the DPI theorem, the information density in the CNN information pipeline must exhibit a decreasing trend.  This information decreasing trend can also explain the phenomenon observed by \cite{survey-efficient-dnn} that the activation outputs of the latter layers of a CNN are increasingly sparse.  We plot Figure \ref{fig-information-flow} to illustrate our theoretically justified conjecture that the latter macroblocks (defined in Section I) of a CNN contain lower information density than the earlier ones.  This conjecture establishes the foundation of our model-reduction scheme.  

\subsection{Lesion Study: Definitions and Setup}

How can the trend of decreasing information density help reduce the model size
of a CNN?  This section conducts lesion studies to shed the insights.
Our lesion consists of three components: the CIFAR-$10$ dataset, a sequential CNN, 
and our designed one-hot lesion:

\textbf{CIFAR-10 Dataset.}
CIFAR-$10$ consists of natural images with resolution $32 \times 32$ of $10$ classes, and divides $50$k images for training and $10$k images for testing.
CIFAR-$10$ is not too large (convenient for us to
run many experiments) and contains sufficient natural
images to reflect CNN behaviors.
The network on CIFAR-$10$ was trained using SGD, weight decay $10^{-4}$, momentum $0.9$, and mini-batch size $128$.
Following the work of \cite{resnet}, we set
the initial learning rate to be $0.1$ and divided by $10$ 
at $50\%$ and $75\%$ of the total epochs, respectively.

\begin{figure*}[t]

	\begin{center}		
		\subfigure[Conv. depth = {$15$}, {$\mathbf{n_b} = [16, 32, 64]$}, {$h_i(\mathbf{n},1)$}]{\includegraphics[width=0.45\textwidth]{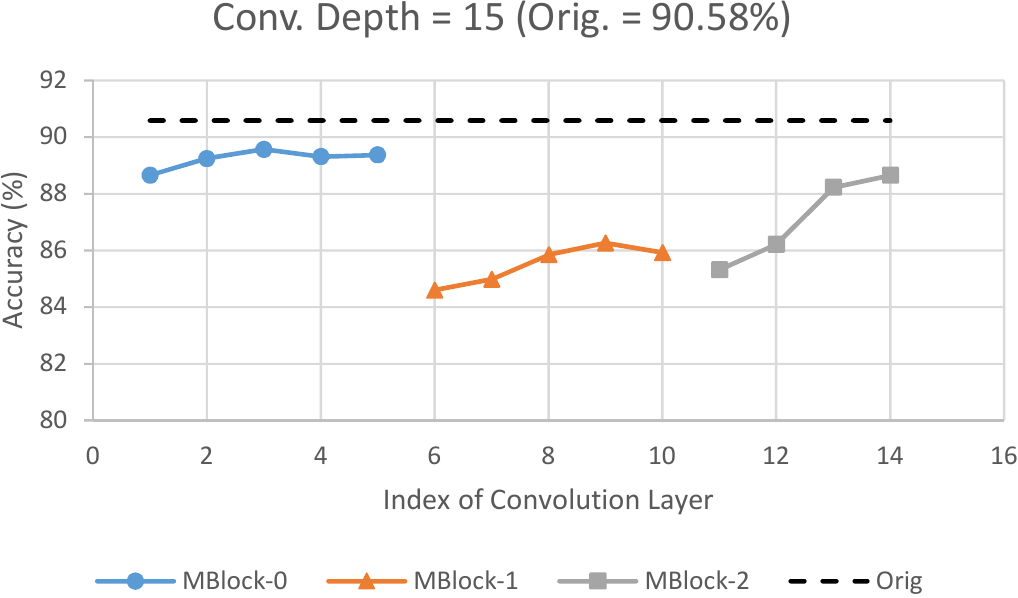}}
		\subfigure[Conv. depth = {$18$}, {$\mathbf{n_b} = [16, 32, 64]$}, {$h_i(\mathbf{n},1)$}]{\includegraphics[width=0.45\textwidth]{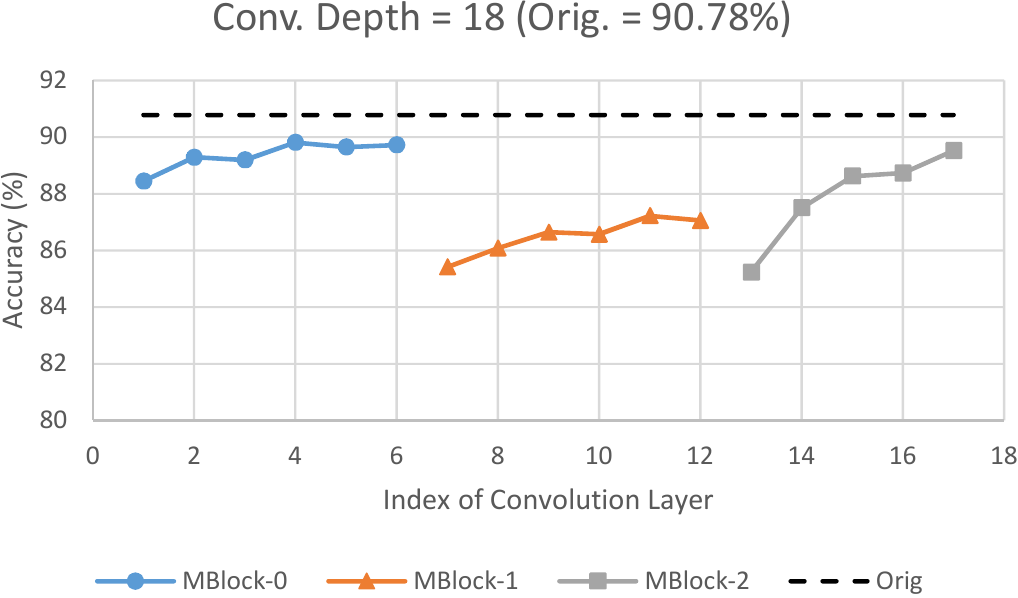}}
		
		\caption{Each point represents a lesion experiment of a one-hot lesion vector $h_i(\mathbf{n},1)$. The dashed black line represents the original prediction accuracies that are (a) $90.58\%$ and (b) $90.78\%$. }
\label{fig-analysis}
\end{center}
\end{figure*}

\textbf{Sequential CNN.}
The sequential CNN consists of the VGG-like structure where the cascaded FC layers are replaced by a global average pooling layer and an FC layer in order to simulate a convolution-layer dominant CNN.
This model serves as our reference model because of its simplicity and representative structures.
In particular, the ``down-sampling then $2 \times$ channel number'' design is widely adopted in modern compact CNN models (e.g., MobileNet \cite{mobilenet}) for considering the trade-off between model complexity (e.g., size) and prediction accuracy.

Figure \ref{fig-cnn} shows an example of  sequential CNNs, which has $15$ convolution layers.
The notation $n_i$ ($i \geq 1$) represents the number of the output channels of convolution layer $L_i$.
We define $n_0$ as the channel number of input images and use the following vector to represent the channels in the sequential CNN,
\begin{equation}\label{eq-w}
\mathbf{n} = [n_0, n_1, \cdots, n_N], \forall \ n_i \in \mathbb{N}.
\end{equation}
The channels that have the same size of output feature maps are grouped into the same \textit{convolution macroblock} $b_i$.
For example, the channels $[n_1, \cdots, n_5]$ ($n_0$ is not an output channel number) in Figure \ref{fig-cnn} are grouped into macroblock $b_0$.
The configuration $\mathbf{n_b}$ of macroblocks of a sequential CNN model is denoted by $\mathbf{n_b} = [n_{b_0}, \cdots, n_{b_m}]$ such as $\mathbf{n_b} = [16, 32, 64]$ for the model in Figure \ref{fig-cnn}.

\textbf{One-hot Lesion.}
We analyze the alteration in the predication accuracy of our sequential CNN models 
by constant and proportional one-hot lesions, depicted as follows:
\begin{itemize}
\item Constant one-hot lesion. For each individual lesion experiment, we set $n_i=c$ for only one selected index $i$ and keep 
the other channel numbers $n_j$ ($j\neq i$) unchanged.
This constant one-hot lesion is denoted as $h_i(\mathbf{n}, c)$. 
\item Proportional one-hot lesion.
For each individual lesion experiment, we set $n_i=k \times n_i$ for only one selected index $i$ and keep the other channel numbers $n_j$ ($j\neq i$) unchanged. Parameter $k$ is tunable. 
This proportional one-hot lesion is denoted as $h_i(\mathbf{n}, k \times n_i)$. 
\end{itemize}

\begin{table}[t]
\scriptsize
	\caption{An example of constant one-hot lesion $h_i(\mathbf{n}, c)$ and proportional one-hot lesion $h_i(\mathbf{n}, k \times n_i)$ for the sequential CNN with convolution depth = $15$}
	\label{tab-lesion}
	\begin{center}
		\begin{tabular}{cc}
			\toprule
			Lesion  & Channel Width (Conv. Depth=$15$) \\ 				\midrule
            Constant & \\
            \midrule
			$\mathbf{n}$ & $[3, 16, 16, 16, 16, 16, 32, 32, 32, 32, 32, 64, 64, 64, 64, 64]$ \\
			$h_1(\mathbf{n},c)$ & $[3, \ c, 16, 16, 16, 16, 32, 32, 32, 32, 32, 64, 64, 64, 64, 64]$  \\
			$\cdots$ & $\cdots$  \\ 
			$h_{14}(\mathbf{n},c)$ & $[3, 16, 16, 16, 16, 16, 32, 32, 32, 32, 32, 64, 64, 64, \ c, 64]$  \\ 
            \midrule
            \midrule
            Proportional & \\
            \midrule
            $\mathbf{n}$ & $[3, 16, 16, 16, 16, 16, 32, 32, 32, 32, 32, 64, 64, 64, 64, 64]$ \\
			$h_1(\mathbf{n},k n_{1})$ & $[3, \ k n_{1}, 16, 16, 16, 16, 32, 32, 32, 32, 32, 64, 64, 64, 64, 64]$  \\
			$\cdots$ & $\cdots$  \\ 
			$h_{14}(\mathbf{n}, k n_{14})$ & $[3, 16, 16, 16, 16, 16, 32, 32, 32, 32, 32, 64, 64, 64, \ k n_{14}, 64]$  \\ 
			\bottomrule
		\end{tabular}
	\end{center}
\end{table}

Table \ref{tab-lesion} illustrates an example of a one-hot lesion for a sequential CNN with convolution depth $15$.

We conduct a one-hot lesion (between convolution layers) first to investigate the prediction accuracy behaviors at the channel level of a sequential CNN model.
We then investigate at the macroblock level about rate distortion (RD) performance, which is an important criterion in lossy compression from the perspective of information theory.

\subsection{Lesion Study: Analysis}

\begin{figure*}[t]

	\begin{center}	
    \subfigure[Conv. depth = {$15$}, {$\mathbf{n_b} = [16, 32, 64]$}, {$h_i(\mathbf{n},\frac{n_i}{16})$}]{\includegraphics[width=0.45\textwidth]{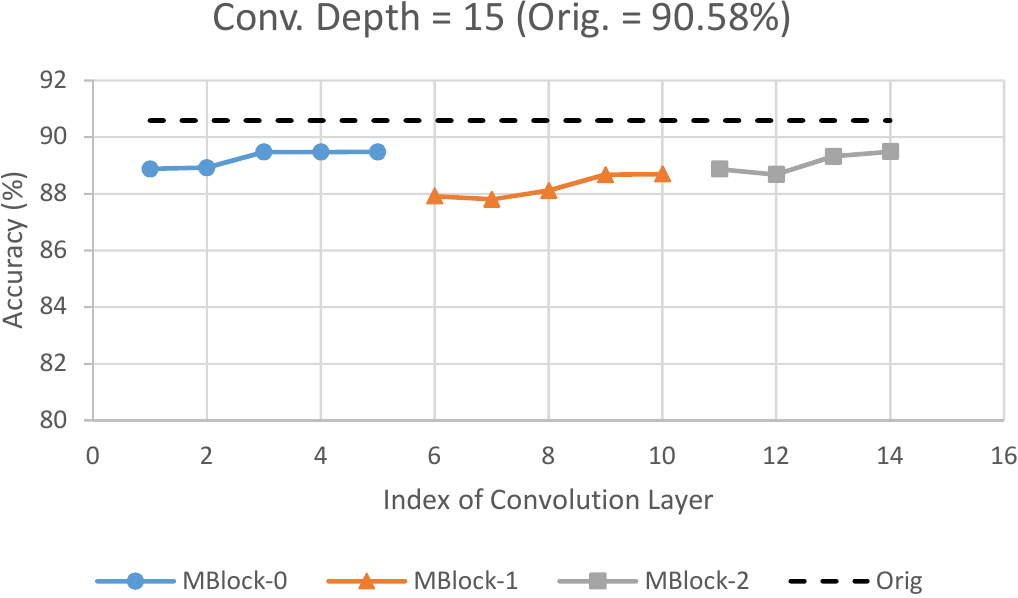}}
		\subfigure[Conv. depth = {$18$}, {$\mathbf{n_b} = [16, 32, 64]$}, {$h_i(\mathbf{n},\frac{n_i}{16})$}]{\includegraphics[width=0.45\textwidth]{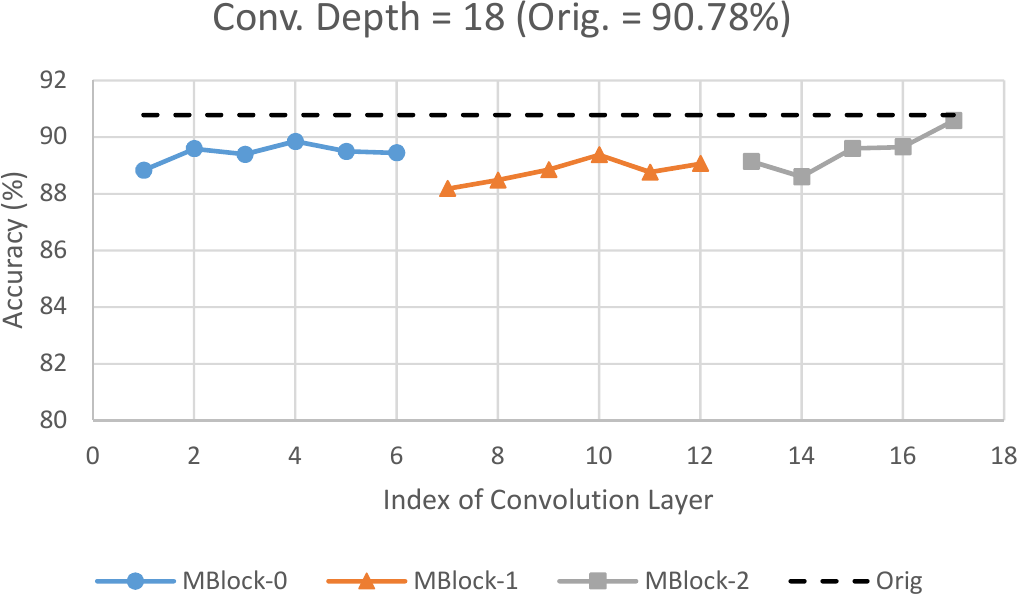}}
        
        \subfigure[Conv. depth = {$15$}, {$\mathbf{n_b} = [16, 32, 64]$}, $h_i(\mathbf{n},\frac{n_i}{8})$]{\includegraphics[width=0.45\textwidth]{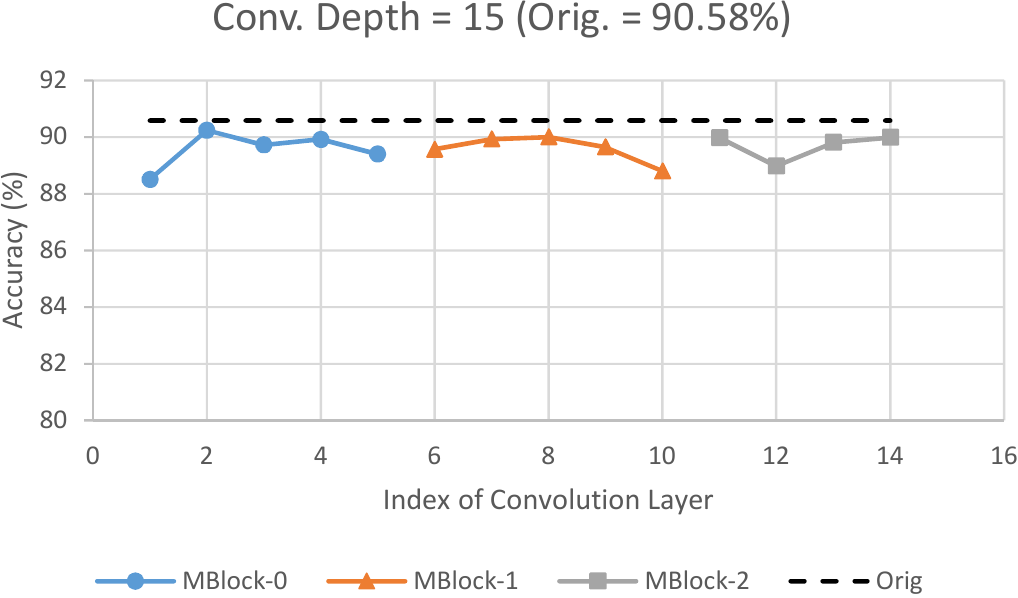}}
		\subfigure[Conv. depth = {$18$}, {$\mathbf{n_b} = [16, 32, 64]$}, {$h_i(\mathbf{n},\frac{n_i}{8})$}]{\includegraphics[width=0.45\textwidth]{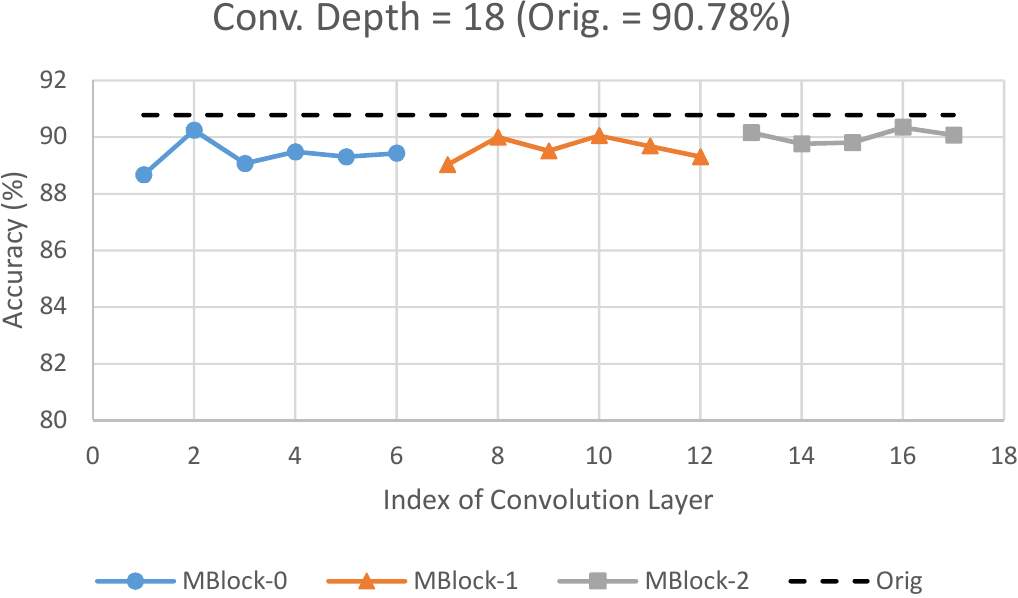}}
		
		\caption{Each point represents a lesion experiment of a proportional one-hot lesion vector. The configuration [conv. depth, one-hot lesion] for each lesion study is (a) [15, {$h_i(\mathbf{n},\frac{n_i}{16})$}], (b) [18, {$h_i(\mathbf{n},\frac{n_i}{16})$}], (c) [15, {$h_i(\mathbf{n},\frac{n_i}{8})$}] and (d) [18, {$h_i(\mathbf{n},\frac{n_i}{8})$}].}
        
\label{fig-analysis-proportional}
\end{center}
\end{figure*}

\begin{figure*}[t]

	\begin{center}		
		\subfigure[Conv. depth = {$15$}, {$\mathbf{n_b} = [16, 32, 64]$}]{\includegraphics[width=0.45\textwidth]{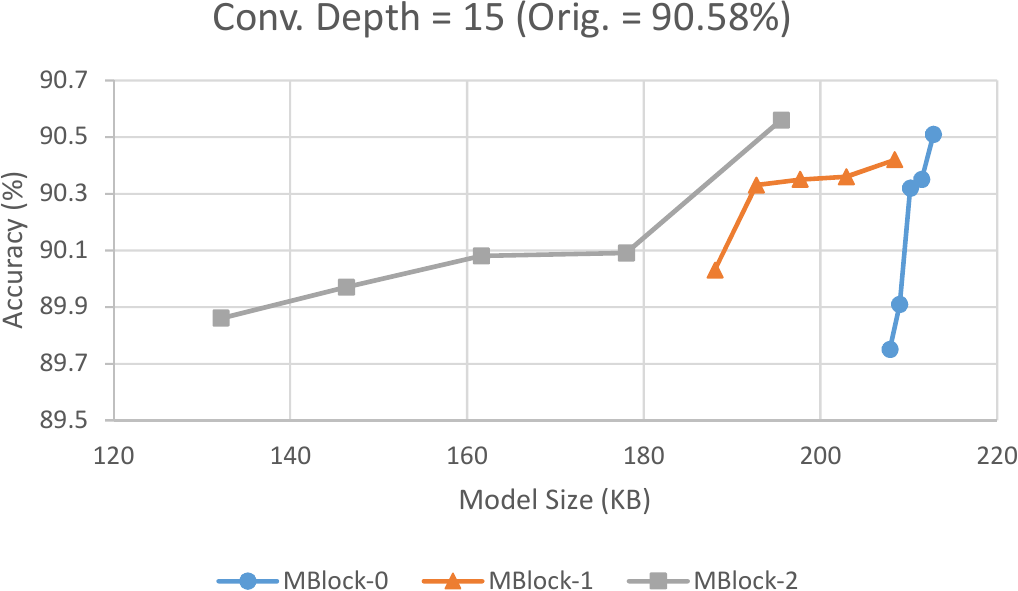}}
		\subfigure[Conv. depth = {$18$}, {$\mathbf{n_b} = [16, 32, 64]$}]{\includegraphics[width=0.45\textwidth]{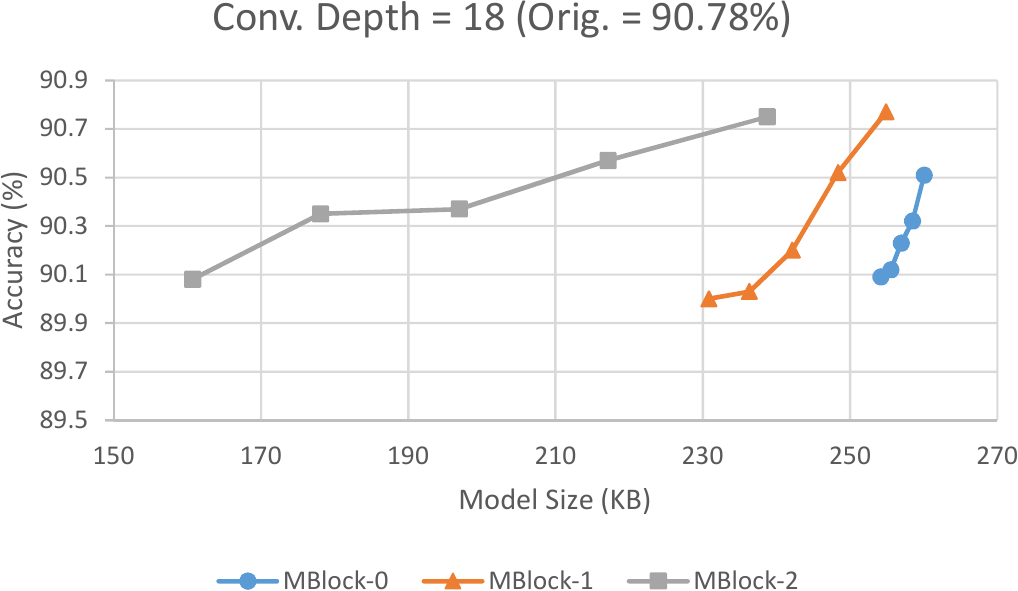}}
		
		\caption{RD curve of each macroblock for $h_i(\mathbf{n_b}, k n_{b_i})$, where $k=[\frac{11}{16}, \cdots, \frac{15}{16}]$. }
\label{fig-analysis-mb}
\end{center}
\end{figure*}

Our experiments aim to find the relationship between model size and model prediction accuracy at different levels: first
at the channel level and then at the macroblock
level. 

In constant one-hot lesion at the channel level, we expected that  
keeping only a constant $c$ channel number in the latter 
macroblocks would suffer from severer degradation
in prediction accuracy.
This expectation is based on the fact that the channels in the
latter macroblocks are much wider, and
 constant one-hot lesion takes more information away
from the channels in the latter macroblocks (e.g., keeping $c$ out of $64$) 
than from the earlier ones (e.g., keeping $c$ out of $32$).
Figure \ref{fig-analysis} plots the result of constant 
one-hot lesion at the channel level.
The x-axis denotes the selected channel index $i$ for 
$h_i(\mathbf{n},c)$.
The y-axis shows the corresponding prediction accuracy of the modified CNN on  CIFAR-$10$.
Surprisingly, macroblock $b_2$ enjoys an unexpected accuracy
\textit{bounce}: the accuracy achieved by removing
more channels from macroblock $b_2$ is higher than the accuracy
achieved by removing less channels from $b_1$.

Figure \ref{fig-analysis-proportional} plots the experimental results of proportional one-hot lesion. Why did we perform this study?  
Constant one-hot lesion keeps only one channel (or $c$ channels) 
alive in each macroblock.  
One may argue that such an extreme study may have more 
severely penalized the later macroblocks with 
larger number of channels being removed.  
Thus, we conduct proportional one-hot lesion to keep alive the same percentage of channels in all macroblocks. 
In the first experiment we set the alive channels to be one sixteenth, and the second experiment one eighth.
Figures \ref{fig-analysis-proportional}(a) 
and \ref{fig-analysis-proportional}(b) 
show that when one sixteenth channels are kept, the 
\textit{bounced} behavior on macroblock $b_2$ still exhibits. 
Since we increase the channel number of proportional one-hot 
lesion, the prediction accuracy gap is reduced 
between macroblocks $b_1$ and $b_2$.
When we further increase the alive channels to one eighth,
Figures \ref{fig-analysis-proportional}(c) 
and \ref{fig-analysis-proportional}(d) show that the 
distortion of all one-hot lesion is limited 
(i.e., $\leq 2\%$ accuracy drop).
The \textit{bounced} behavior still
exists thought less significant.
Figure \ref{fig-analysis-proportional}(d) illustrates 
that macroblock $b_2$ has the highest average prediction 
accuracy among all macroblocks.

These two lesion studies demonstrate that macroblock $b_2$ provides better accuracy recovery as compared with macroblock $b_1$ even though we remove much more channels from $b_2$ in both constant and proportional one-hot lesion studies.
Two observations can be made from these two studies. 
First, CNN is robust
for information-loss recovery.  Second, the channels in the latter
macroblocks seem to have higher degrees of information redundancy.  

We next extend proportional one-hot lesion to the macroblock level to 
investigate the relation between model size and prediction accuracy.
Let $h_i(\mathbf{n_b}, k \times n_{b_i})$ denote channel number reduction in 
the macroblock level, which we reduce all channels in macroblock $b_i$ from $n_{b_i}$ to $k \times n_{b_i}$ ($0 < k < 1$) and keep the other channels 
in $b_j$ ($j \neq i$) intact.

Figure \ref{fig-analysis-mb} shows the RD performance of each macroblock by $h_i(\mathbf{n_b}, k \times n_{b_i})$, 
where $k=[\frac{11}{16}, \cdots, \frac{15}{16}]$.
The x-axis depicts the model size of the reduced CNN model 
from small on the left-hand side to large on the right-hand side.
The y-axis depicts the corresponding prediction accuracy of the 
modified CNN trained on CIFAR-$10$.
Figure \ref{fig-analysis-mb} shows 
that macroblock $b_2$ (in gray) enjoys the best RD performance 
(by maintaining high prediction accuracy) among the macroblocks.
In addition, macroblock $b_2$ enjoys the best compaction
since the model size is significantly reduced by compressing macroblock $b_2$ as compared with compacting the other macroblocks.

These lesion studies lead to our following hypotheses: 

\begin{itemize}
\item {\em The latter macroblock contains lower information density.} 
We can observe that the latter macroblock has lower information density based on our lesion studies.
The last macroblock $b_2$ has better prediction accuracy even after we have removed more channels from it than from macroblock $b_1$.

\item {\em Network distortion is based on information density.}
The actual network distortion is correlated with the information density instead of simply the number of removed channels.
Therefore, the network demonstrates strong capabilities to recover from the distortions when we remove the channels with low information density.
\end{itemize}

\subsection{Backward Reduction for Model Reduction}
\label{sec-proposal}

\begin{figure}[t]
	\vskip 0.2in
	\begin{center}
		\centerline{\includegraphics[width=0.9\columnwidth]{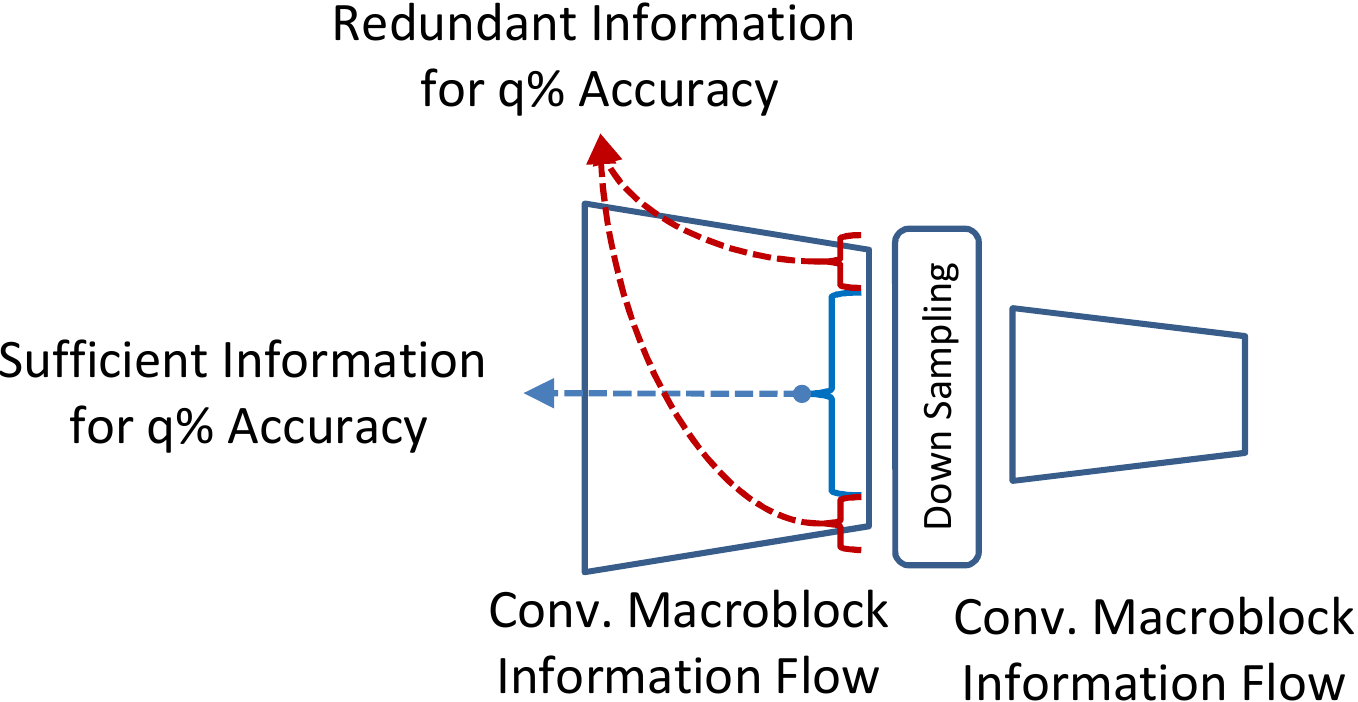}}
		\caption{A conceptual illustration of the proposed backward reduction method that is based on the conjectured information flow structure.}
		\label{fig-backward-reduction}
	\end{center}
\end{figure}

Figure \ref{fig-backward-reduction} describes the key idea of 
our proposed backward reduction method.  The down-sampling layer plays 
the role of 
being an information gateway, through which information 
(some may be redundant) passes  
for achieving $q\%$ prediction accuracy.
We can reduce the information flow by decreasing the channel number of the previous convolution macroblock as long as there maintains sufficient information for achieving the target accuracy.
While applying this idea recursively, we must ensure the information to
be still sufficient for the classification task in the FC-Layer.
Since the classification is performed at the end of a CNN pipeline, 
our model reduction process starts from the last convolution macroblock.
Another reason to perform backward instead of forward 
reduction is that our lesion studies reveal the latter macroblocks 
containing lower information density. We will shortly justify
this backward choice by presenting experimental results.

\begin{algorithm}[t] 
\caption{Backward Reduction} 
\label{alg-backward-reduction} 
\begin{algorithmic}[1]
	\REQUIRE  $[b_0,\cdots,b_m]$ and $\delta$\\
    /* $[b_0,\cdots,b_m]$: Convolution macroblocks \\
    /* $\delta$: Threshold of accuracy distortion 
    \ENSURE $\beta = [\beta_0, \beta_1, \cdots, \beta_m]$ \\
    /* $\beta$: Macroblock-wise Channel Width Multipliers 
	\FOR{macroblock $b_i :=b_m$ to $b_0$} 
      \STATE $U = 1, L=0.5$
      \WHILE{$[(U-L)n_{b_i}] > 1$}
          \STATE $\beta_i' = (L+U)/2$
          \STATE $n_{b_i}' = \left \lceil \beta_i' n_{b_i} \right \rceil$, retrain CNN
          \IF{Accuracy distortion $< \delta$}
              \STATE $U = \beta_i'$
          \ELSE{} 
          	\STATE $L = \beta_i'$
          \ENDIF
      \ENDWHILE
      \STATE $\beta_i = \beta_i'$
    \ENDFOR
    \STATE return $\beta = [\beta_0, \beta_1, \cdots, \beta_m]$
\end{algorithmic}
\end{algorithm}

Algorithm \ref{alg-backward-reduction} presents our proposed
\textit{backward reduction} algorithm (named BRIEF) based on the information flow structure conjectured from a combination of information flow density and the dynamic nature of CNNs.

The model reduction problem of a CNN can be formulated as follows
\begin{equation}\label{eq-reduction}
\min_{\mathbf{n'}} R(\mathbf{n'}), \textup{  subject to } D(\mathbf{n'}, \mathbf{n}) \leq \delta,
\end{equation}
where $R(\mathbf{n'})$ represents the bit rate of the model for the optimized channel width configuration $\mathbf{n'}$, $D(\mathbf{n'}, \mathbf{n})$ denotes the prediction accuracy degradation as compared with the original performance, and $\delta$ is the acceptable performance loss budget.

The model reduction formulation of Eq. (\ref{eq-reduction}) encapsulates an integer programming problem (i.e., channel number $n_{i}$'s must be positive integers), which makes the optimization an NP-hard problem.
Furthermore, no closed-form distortion evaluation function exists, since the CNN is usually considered to be a black-box.
Fortunately, with the aid of the conjectured information flow structure, we can reduce the original NP-hard problem to solving a one-dimensional greedy search problem. 
We search for the proper scaling factor $\beta_i$ staring from the last convolution macroblock,
\begin{equation}\label{eq-theta}
\min_{\beta_i} n_{b_i}' = \lceil \beta_i n_{b_i} \rceil, \textup{  subject to } D(\mathbf{n}, \mathbf{n'}) \leq \delta.
\end{equation}
That is, we optimize for the proper value of the scaling factor $\beta_m$ for the last convolution macroblock $b_m$ within distortion budget $\delta$. 
Then we iterate through the same greedy search process for the previous convolution macroblock $b_{m-1}$.

Our algorithm optimizes for the channel width 
multiplier $\beta_i$ for each convolution macroblock from $b_m$ down to $b_0$.
The initial setting of the lower bound $L=0.5$ in step $2$ is based on the observation by MobileNet \cite{mobilenet} that
there is a significant drop in prediction accuracy 
when the channel multiplier is less than $0.5$.
The default distortion budget $\delta$ in our algorithm is set to $1\%$. 

Our algorithm BRIEF retrains the CNN from scratch (step $5$) to investigate 
the prediction accuracy of current configuration since our lesion studies have demonstrated network robustness against channel removal.
The retraining process can be relaxed by fewer training epochs as long as the training setting can retain the original CNN prediction accuracy.
Once the accuracy distortion is acquired, we can adjust the upper bound and the lower bound for  $\beta_i$ accordingly using 
a greedy binary search method (steps $7$ and $9$).

Earlier we mentioned that to ensure information being sufficiently
maintained for the classification task, which is performed at the last layer 
of a CNN, we propose performing backward reduction.  
To confirm this backward heuristic to be accurate, we
conducted experiments to compare the effectiveness of backward versus forward reduction.
Table \ref{tab-reduction-seq} shows that backward reduction outperforms forward reduction significantly in size ($45.3\%$ versus $6.3\%$) at a similar prediction accuracy.
As a result, the backward reduction algorithm acts to remove the channels with low information density.
Therefore, the network can recover information loss from the latter macroblocks when there is sufficient information provided from the earlier macroblocks.
On the other hand, the forward reduction approach removes 
information starting from the input source, 
which distorts the original information and 
eliminates the network's ability to recover from such distortion.

\begin{table}[t]
	\caption{Performance comparison between the forward reduction (reduction sequence $[b_0,b_1,b_2]$) and the backward reduction (reduction sequence $[b_2, b_1, b_0]$) for the sequential CNN with convolution depth = $15$ on CIFAR-$10$.}
	\label{tab-reduction-seq}
	\begin{center}
		\begin{tabular}{clcc}
			\toprule
			Model & Acc. (\%) [Diff.] & Size (MB) & Saving (\%) \\ \midrule \midrule
			Original & $91.24$ & $0.87$ & - \\ 
            Forward & $90.46$ $[0.78]$ & $0.82$ & $6.3$ \\ 
            Backward & $90.31$ $[0.93]$ & $0.48$ & $\mathbf{45.3}$ \\ 
			\bottomrule
		\end{tabular}
	\end{center}
\end{table}

\section{Experimental Results}
This section reports experimental results with BRIEF on the ImageNet dataset for various CNN models.
We conducted BRIEF's backward reduction algorithm on the last two convolution macroblocks.
We implemented BRIEF with PyTorch $0.3$. 
Our evaluation metric is model size.
(We care about model compaction not to significantly degrade prediction 
accuracy, but we do not compare accuracy between different CNN models.)
To realistically measure model size, we use the
actual required storage 
(including headers) of a trained model (i.e., state\_dic in PyTorch).

\subsection{ImageNet Dataset}

ImageNet consists of $1.28$M training images and $50$k validation images, divided
into $1,000$ classes.
We trained our models for $90$ epochs with a batch size of $256$. 
The learning rate, set to $0.1$ initially, is divided by $10$ at epoch $30$ and $60$.
Simple data augmentation was adopted based on the ImageNet script by PyTorch, which is the same as \cite{resnet}. 
The single-center-crop validation accuracy is reported for the CNN models.

In order to reduce the computational effort of BRIEF's 
greedy backward reduction process on ImageNet, we empirically adopted the shorter training setting of $20$ training epochs with a learning rate (set to $0.1$ initially) divided by $10$ at epoch $8$ and $16$.
We used $20$-epoch setting in Algorithm \ref{alg-backward-reduction} to quickly find the best $\beta$, and eventually used this $\beta$ to train the final model for $90$ epochs.
Then, we evaluated the final performance by training the CNN in the original $90$-epoch training setting.

For the shorter training setting, we designed and performed experiments to determine training epochs. 
In these experiments, we trained several CNN models for different training epochs and fixed the other hyper-parameters of the training. 
The different choices of training epochs we explored are $10$, $20$ and $30$. 
We observed that 1) there is considerable discrepancy between the results of $10$-epoch and $90$-epoch settings, 2) the results of $20$-epoch setting can most reflect the results of $90$-epoch setting, and 3) the results of $20$-epoch and $30$-epoch settings are basically the same. 
Based on these three observations of our designed experiments, we thus selected $20$ as our shorter training setting.

\begin{table}[t]
	\caption{Performance comparison between our method and the previous work of ResNet-$34$ on ImageNet.}
	\label{tab-compare}
	\begin{center}
		\begin{tabular}{llcl}
			\toprule
			ResNet-$34$ & Top-$1$ (\%)  & Param. ($10^7$) & Saving  \\
			\midrule
			\midrule
			Baseline of \cite{prune-filter} & $73.23$ & $2.16$ & - \\ 
			\cite{prune-filter} & $72.17$ $[1.06]$ & $1.93$ & $10.8$\% \\ \midrule
			Baseline & $\mathbf{73.54}$ & $2.18$ & - \\
			Proposed & $\mathbf{72.49}$ $[\mathbf{1.05}]$ & $\mathbf{1.48}$ &  $\mathbf{32.3\%}$ \\		
			\bottomrule
		\end{tabular}
	\end{center}
\end{table}

\begin{table*}[t]
	\caption{Backward reduction results for the CNN models. The configuration shows the channels numbers of the last two convolution macroblocks. We report the scaling factors for SqueezeNet due to the irregular structure.}
	\label{tab-imagenet}
	\begin{center}
		\begin{tabular}{llcccr}
			\toprule
			Model & Top-$5$ (\%) [Diff.] & Param. ($10^6$) & Size (MB) & Saving (\%)  & Config. \\
			\midrule
			\midrule
			ResNet-$34$ & $91.410$ & $21.80$ & $83.24$ & - & $[256, 512]$ \\
			Proposed & $90.854$ $[0.556]$ & $14.80$ & $56.52$ & $\mathbf{32.10}$ & $[256, 346]$ \\
			\midrule
			ResNet-$18$  & $89.350$ & $11.69$ & $44.64$ & - & $[256, 512]$ \\
			Proposed & $88.866$ $[0.484]$ & $8.45$ & $32.28$ & $\mathbf{27.69}$ & $[245, 405]$ \\
			\midrule
			$1.0$-MobileNet  & $88.520$ & $4.23$ & $16.25$ & - &  $[512, 1024]$  \\
			Proposed & $87.952$ $[0.568]$ & $2.64$ & $10.15$ & $\mathbf{37.56}$ &  $[507, 513]$ \\
			\midrule
			$0.75$-MobileNet  & $86.992$ & $2.59$ & $9.95$ & - & $[384, 768]$ \\
			Proposed & $86.500$ $[0.492]$ & $1.67$ & $6.44$ & $\mathbf{35.28}$ & $[369, 442]$ \\
			\midrule
			$0.5$-MobileNet  & $83.972$ & $1.33$ & $5.14$ & - & $[256, 512]$ \\
			Proposed & $82.792$ $[1.180]$ & $0.94$ & $3.63$ & $\mathbf{29.31}$ & $[252, 331]$ \\
			\midrule
			SqueezeNet $1.1$ & $79.220$ & $1.24$ & $4.72$ & - & $\beta=[1.0, 1.0]$ \\
			Proposed & $78.602$ $[0.618]$ & $1.10$ & $4.21$ & $\mathbf{10.81}$ & $\beta=[0.8, 0.9]$ \\
			\bottomrule
		\end{tabular}
	\end{center}
\end{table*}

\subsection{Evaluation of Various CNN Models}
Table \ref{tab-compare} reports that BRIEF significantly outperforms the state-of-the-art approach \cite{prune-filter} conducted on ResNet-$34$, a convolution-layer dominant CNN model.
BRIEF reduces the model by $3 \times$ more ($32.8\%$) than the previous work does ($10.8\%$) with similar top-$1$ accuracy.
Our algorithm removed the channels aggressively even including the non-zero weighting parameter as long as the prediction accuracy is maintained.
Therefore, we can explore the additional regions that are ignored by the traditional approaches.
This result confirms the great potential in model 
reduction from the perspective on information flow.

We evaluated our method on a variety of Convolution-layer dominant CNN models (e.g., ResNet, MobileNet, and SqueezeNet) since these compact models are more challenging targets.
MobileNet is a tough reduction target because it has a width-multiplier technique that adjusts all the channel numbers by the same scaling factor.
We also evaluated the proposed backward reduction method on the further 
scaling-reduced MobileNets (e.g., $0.5$-MobileNet).
We reported the scaling factors for the final two macroblocks of SqueezeNet with irregular structure and the actual channel number in the experiments for the CNNs with regular structures.

Table \ref{tab-imagenet} presents the performance of BRIEF among the various Convolution-layer dominant CNN models.
The configuration column of Table \ref{tab-imagenet} shows the architecture of each convolution macroblock before and after the proposed backward reduction algorithm.
The accuracies of the CNN models in Table \ref{tab-imagenet} are evaluated using the $90$-epoch training setting.
We achieved an up to $37.56\%$ model size reduction 
on MobileNet with only a $0.56\%$ accuracy loss.

Notice that the model reduction ratio decreases as we applied
BRIEF on the further optimized MobileNets (e.g., $0.75$-MobileNet, $0.5$-MobileNet).
Even for $0.5$-MobileNet, we still accomplished a $29.3\%$ model size reduction, which resulted in a storage size of only $3.63$ MB.
The model size of the reduced $0.5$-MobileNet ($3.63$ MB) is already smaller than that of SqueezeNet ($4.72$ MB), while the prediction accuracy of MobileNet still outperforms SqueezeNet.
For the highly optimized CNN, SqueezeNet, we achieved a $10.81\%$ model reduction while the accuracy loss remains within $1\%$.

\begin{figure}[t]
	\begin{center}
		\centerline{\includegraphics[width=0.95\columnwidth]{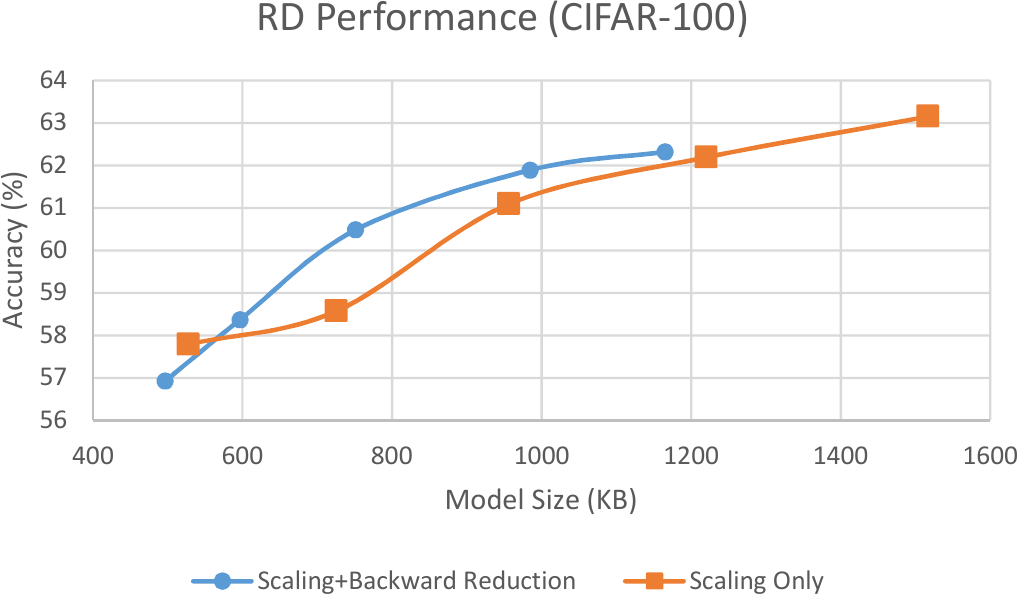}}
		\caption{Rate-Distortion curve of the scaling-only and the scaling with backward reduction approaches.}
		\label{fig-rd-curve}
	\end{center}
\end{figure}

\subsection{Rate-Distortion Behavior}

We compared our backward reduction algorithm to the traditional $\alpha$-scaling approach that multiplies the whole CNN to the same scaling factor $\alpha$.
The configuration is set to $\mathbf{n_b} = [32, 64, 128]$ for the sequential CNN on CIFAR-$100$ in this experiment.
Figure \ref{fig-rd-curve} shows the RD curves of the scaling-only approach and the scaling with our backward reduction algorithm.
The scaling factor $\alpha$ ranges from $0.5$ to $0.9$ with step size $0.1$. The $x$-axis represents the required bit rate (KB), and the $y$-axis denotes the prediction accuracy. The RD performance experiment is conducted on the sequential CNN with convolution depth = $12$.
For the vast majority of models, the $\alpha$-scaling with BRIEF 
further improves the simple $\alpha$-scaling approach in terms of RD performance.

\section{Conclusion}

The conjectured information flow structure describes CNNs as dynamic structures rather than static ones. 
As long as sufficient information is provided from the prior convolution macroblocks, we can safely remove the redundant channels and still maintain the same level of accuracy.
Therefore, we can explore the additional regions of CNNs, which were not possible with the traditional filter magnitude based approaches.
Using our backward reduction algorithm BRIEF, we reduced ResNet-$34$ significantly, attaining $3 \times$ better reduction than previous approaches.
We also reduced MobileNet to just $3.63$ MB, making it smaller than SqueezeNet ($4.72$ MB), while achieving a slightly higher prediction accuracy.
The capability of backward reduction was validated on highly compact CNNs, including SqueezeNet and MobileNet.

Our method is applied at the macroblock level, which the greedy algorithm may impose the computational burden for searching the proper channel number for the macroblock.
We plan to derive the minimum required channel number directly with the aid of further information flow analysis for model reduction in our future work.
The proposed greedy algorithm demonstrates the potential of model reduction by leveraging the information flow analysis, which we expect this work to act as a stepping stone towards opening the black-box of CNNs to establish more efficient and compact CNN models.
Our macroblock level model reduction method complements to the reduction approaches conducted in the fine-grained level (e.g., sparse filter connection), which will achieve a compound improvement in model compression.

\bibliographystyle{IEEEtran}

\end{document}